# Automatic Acrostic Couplet Generation with Three-Stage Neural Network Pipelines

Haoshen Fan[1,2], Jie Wang[2], Bojin Zhuang[2], Shaojun Wang[2] and Jing Xiao[2]

[1] University of Science and Technology of China
[2] Ping An Technology (Shenzhen) Co., Ltd, China
sa517069@mail.ustc.edu.cn, photonicsjay@163.com,
zhuangbojin232@pingan.com.cn, wangshaojun851@pingan.com.cn,
xiaojing661@pingan.com.cn

**Abstract.** As one of the quintessence of Chinese traditional culture, couplet compromises two syntactically symmetric clauses equal in length, namely, an antecedent and subsequent clause. Moreover, corresponding characters and phrases at the same position of the two clauses are paired with each other under certain constraints of semantic and/or syntactic relatedness. Automatic couplet generation is recognized as a challenging problem even in the Artificial Intelligence field. In this paper, we comprehensively study on automatic generation of acrostic couplet with the first characters defined by users. The complete couplet generation is mainly divided into three stages, that is, antecedent clause generation pipeline, subsequent clause generation pipeline and clause re-ranker. To realize semantic and/or syntactic relatedness between two clauses, attention-based Sequence-to-Sequence (S2S) neural network is employed. Moreover, to provide diverse couplet candidates for re-ranking, a cluster-based beam search approach is incorporated into the S2S network. Both BLEU metrics and human judgments have demonstrated the effectiveness of our proposed method. Eventually, a mini-program based on this generation system is developed and deployed on Wechat for real users.

## 1 Introduction

Chinese antithetical couplet, (namely "对联"), which consists of two clauses, is an important part of Chinese cultural heritage. As a part of Chinese people's cultural life, couplets have become a popular way to expressing personal emotion, political views, or communicating blessing messages at festive occasions. As an important traditional cultural game, given one antecedent clause, people are challenged to write the subsequent clause. Additionally, couplets expressing blessing and happiness are written on red banners on special days, such as the Chinese New Year, birthday and wedding ceremonies. Literally, Chinese couplet must satisfy certain constraints on syntactic and/or semantic relevance. For example, corresponding characters or phrases from the same positon in the two clauses must be paired with each other. For instance, as shown in **Fig. 1**, the character 'hundred' is paired with 'thousand', 'flower' is antithetical to 'tree', 'good' correspond to 'new' and 'bloom' is coupled with 'boom'. Compared to common proses such as news and fictions, couplet also exhibits poetic aestheticism, e.g., rhyming and conciseness etc.

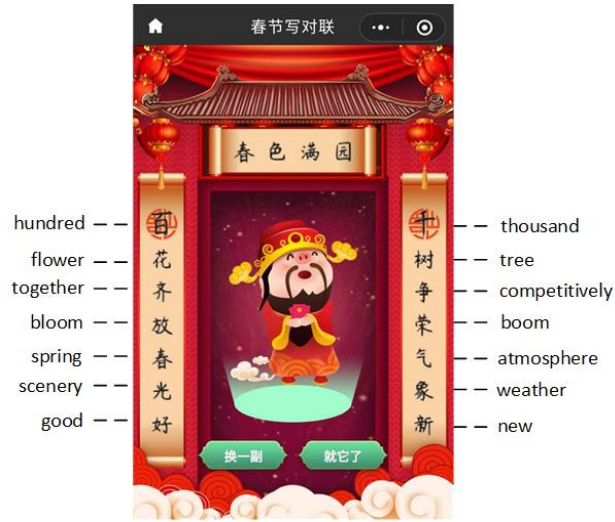

**Fig. 1.** An acrostic couplet generated by our developed mini-program of couplet generation. Each Chinese character is translated into English for reference. The abstract meaning of this couplet is that many flowers blossom in spring making the spring scenery very beautiful (antecedent clause, left); many trees competitively boom in spring which refresh the atmosphere (subsequent clause, right).

In this paper, we focus on automatic acrostic couplet generation with deep learning methods. Especially at Spring Festival, couplet have been popularly used for expressing blessing for the coming new year. Under the analysis of potential user demand, we had planned to develop an online mini-program of automatic couplet generation on Wechat. Contrary to the couplet game that people are challenged to write subsequent clause given the antecedent one, our automatic acrostic couplet generation can compose a complete clause pair with users' intent defined in both clause heads, which facilitates fluent user interaction. Obviously, as opposed to completing subsequent clause, automatic acrostic couplet generation is more challenging. Herein, we formulate the acrostic couplet generation as a three-stage natural language generation problem. In the first stage, the antecedent clause is generated by a pipeline of recurrent neural network based language model (RNN-LM) given user' intent as head characters. Afterwards, the subsequent clause is generated by an attention-based S2S network by taking in the antecedent one from the previous stage. Moreover, a cluster-based beam search (CBS) method is incorporated to generate a candidate pool of diverse couplets. Eventually, best couplet is selected from the candidate pool with a re-rank pipeline.

In order to creating interesting and excellent Spring Festival acrostic couplets for online users, the re-ranking pipeline is based on the following criterions. For example, the length of single clause is in the range from 5 to 12 characters. In addition, corresponding characters at the same positon of two clauses should have the same part of speech (POS). Ending tone of both clauses must be opposed. For instance, if pronounce tone of last character in the antecedent clause is level, the corresponding tone of last character in the subsequent one must be oblique. The rest of this paper is organized as follows. In Section 2, the related work of couplet generation is introduced. The detail of our model is described in Section 3. Section 4 summarizes experimental results and our study is concluded in Section 5.

## 2 Related Work

Natural language generation (NLG) (Mann, 1982), also known as text generation, is one of most important tasks in the field of natural language processing (Chowdhury, 2003). Compared to convolutional neural network (CNN) (Kalchbrenner et al., 2014), recurrent neural network (RNN) (Mikolov et al., 2010) is more suitable for NLG due to its sequential prediction capability. Moreover, RNN with long-short term memory (LSTM) (Hochreither and Schmidhuber, 1997) or gated recurrent unit (GRU) (Cho et al., 2014) can capture longer contextual information. Recently, Sequence-to-Sequence (S2S) (Sutskever et al, 2014) was proposed for heterogeneous data translation. Furthermore, Bahdanau (Bahdanau et al, 2015) proposed the attention mechanism to diffuse decoding weights into different parts of input, which ensures semantic alignment between input and out sequences.

To some extent, couplet generation can be considered as a similar case of statistical machine translation or poetry generation. There are two main methods for machine translation: Statistical Machine Translation (SMT) (Koehn, 2010) and Neural Machine Translation (NLT) (Koehn, 2017). For example, Koehn (Koehn, 2003) proposed an approach of Statistical phrase-based translation and Devlin (Devlin et al, 2014) proposed Neural Network Joint Model (NNJM) which was constructed using the context of both source and target language. Recently, Ahmed (Ahmed et al., 2018) applied the state-of-the-art transformer structure for machine translation. On the other hand, some researchers proposed the methods based on rules or templates, e.g., phrase search approach (Wu et al., 2009), template search approach (Oliveira, 2012) and summarization approach (Yan et al., 2013) for poetry generation. Furthermore, Zhang and Lapata (Zhang and Lapata, 2014) proposed a poetry generation model based on RNN which generates each line character by character. In order to achieve semantic coherence, a novel two-stage poetry generating method (Wang et al., 2016) was presented. In order to create flexible and creative Chinese poetry, Zhang (Zhang et al., 2017) extended the neural model with memory augment, which balanced the requirements of linguistic accordance and aesthetic innovation.

Under most circumstance, the results generated by end-to-end (E2E) system are not guaranteed to be always satisfied. To address this problem, few researchers re-rank generated texts to select desired results. Jiang (Jiang and Zhou, 2008) used multiple features including Mutual information score and MI-based structural similarity score to train a SVM model for candidates re-ranking. Sordoni (Sordoni et al, 2015) applied the ranking algorithms LambdaMART as supervised ranker.

To the best of our knowledge, nevertheless, few research work focused on the task of couplet generation. Zhang (Zhang and Sun, 2009) proposed a couplet generation model based on statistics and rules. Jiang (Jiang and Zhou, 2008) regarded this task as a kind of machine translation and reported a phrase-based statistical machine translation (SMT) approach. Furthermore, Yan (Yan et al., 2016) proposed a novel polishing schema to refine the generated couplets using additional information. However, our study is different from all above methods. Most previous work tended to treat couplet generation as a special translation task and tried to generate subsequent clauses given antecedent ones. However, a complete acrostic couplet can be generated by our method only given few head characters defined by users. Moreover, contrary to previously reported E2E approaches, three pipelines consisting of a RNN-LM, an attention-based S2S and a re-ranker are combined for better couplet generation for real-time online Wechat users. To provide a candidate pool of diverse couplets for selection, a CBS method is incorporated into the S2S network. Furthermore, we believe that our propose three-pipeline generation method can also be extended to other language generation tasks, such as poetry and stories.

## 3   Model

In this paper, our proposed acrostic couplet generation method mainly consists of three stages, namely, antecedent clause generation pipeline, subsequent clause generation pipeline and the re-rank pipeline as shown in **Fig. 2**. Among an acrostic couplet, head characters of both clauses are defined by users, denoting as $K_1$ and $K_2$. Thus, the two clauses are denoted as: $S_1 = \{K_1, C_{1,2}, C_{1,3}, \dots, C_{1,m}\}$ and $S_2 = \{K_2, C_{2,2}, C_{2,3}, \dots, C_{2,m}\}$, where $m$ represents the clause length minus 1. The antecedent clause $S_1$ can be automatically generated by the RNN-LM pipeline given the head $K_1$. Based on the generated clause $S_1$, the subsequent clause $S_2$ can be generated with an attention-based S2S. Moreover, a CBS method is incorporated to realize a candidate pool of diverse clauses. Eventually, a re-ranking pipeline is used to select the best couplet from the candidate pool.

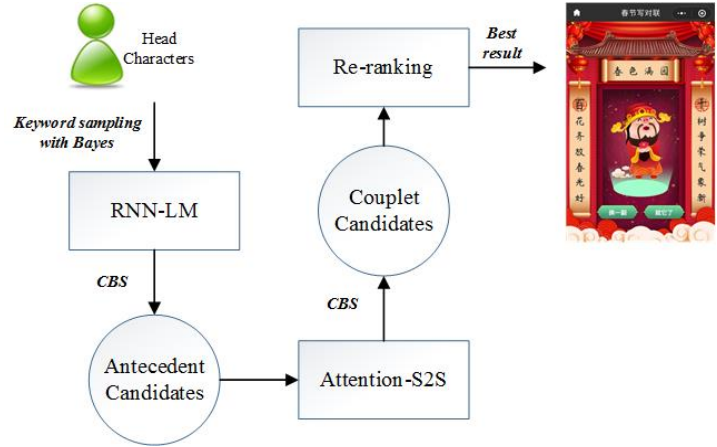

**Fig. 2.** The main framework of our model.

### 3.1 Antecedent Clause Generator

Given the head characters, an RNN-LM model is used to generate antecedent clauses. Neural language model (Bengio et al., 2003) was first proposed in 2003, then Mikolov (Mikolov, 2010) extended it with RNN. We use the vanilla RNN cell to calculate and store the information of each character $C_{1,2}, C_{1,3}, \ldots, C_{1,m}$ in the antecedent clause. Taking in the word embedding of a character $C_{1,i}$, and the previous state $s_{i-1}$, the RNN cell can calculate a current hidden state $s_i$ as follows:

$$s_i = f(w_s s_{i-1} + w_c C_{1,i} + b) \tag{1}$$

where $w$ and $b$ are trainable parameters as weights and bias, and the parametrized non-linear function $f$ is based on hyperbolic tangent. As shown in **Fig. 3**, next character $C_{1,i+1}$ is predicted by the hidden state $s_i$. To generate antecedent sentence diversely and effectively, a CBS method was applied in the decoder, which was proposed by Tam (Tam et al., 2019) to overcome the shortcoming that beam search tends to output several sentences with slight difference. As shown in **Algorithm 1**, CBS combines K-means cluster and beam search to generate more meaningful response. In each decoding step of beam search, CBS perform K-means cluster according to the average embedding of candidates and remove half of candidates in each cluster.

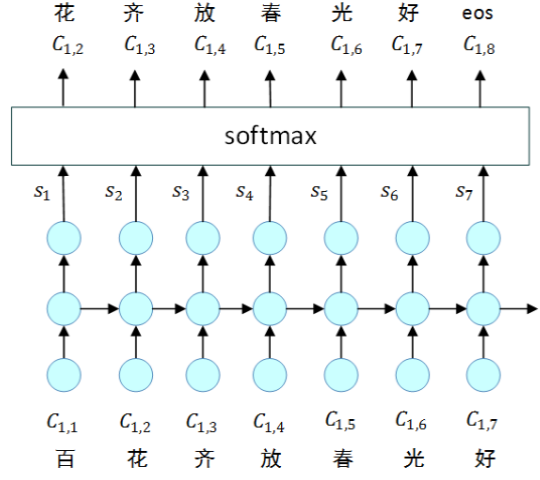

**Fig. 3.** Schematic diagram of the antecedent clause generator.

---

**Algorithm 1:** Cluster-based beam search

**Input:** Beam width *BW*, Candidates *C* initialized with start symbol, maximum decoding step $t_{max}$, cluster number *K*
**Output:** Final result *res*
**While** *Number of completed hypothesis does not reach BW or decoding step not reach* $t_{max}$
**do**
   **for** *i in BW* **do**
      tmpHyps = Top-N(Extend(*C*[*i*], *BS* × 2));
      Remove hyp in tmpHyps with repeated N-grams or UNK;
      Save tmpHyps to extended candidates $C_e$;
   **end**
   Perform K-means over extended candidates $C_e$;
   **for** *candidates in each cluster* **do**
      Sort candidates by partial log-prob scores;
      Choose top *BW/K* candidates;
      Put candidates with end symbol in *R*;
      Put incomplete candidates in *C*;
   **end**
**end**
*res* ← sort *R* according to log-prob scores;

---

### 3.2 Subsequent Clause Generator

The basic idea of subsequent clause generator is to map the antecedent clause into a fix dense vector and then decode the subsequent clause iteratively and sequentially. Sequence-to-sequence (S2S) (Sutskever et al., 2014) is a popular framework for this task. To enhance syntactic and/or semantic relatedness between antecedent and subsequent clauses, attention mechanism is incorporated into the S2S generation model. The generation model iteratively encodes the antecedent clause into a fix dense contextual vector $s_i$. Contrast to the conventional S2S which rely context vector on the last input hidden state $s_m$, attention mechanism considers contribution of each input character into a new context vector as follows:

$$v_t = \sum_{j=1}^{m} a_{tj} s_j \tag{2}$$

The $a_{tj}$ is determined by the previous hidden state $h_{t-1}$ and each hidden state of encoder, i.e., $\{s_1, s_2, \ldots, s_m\}$. Therefore, the new context vector is a weighted sum of hidden states of the encoder, which can adaptively pay attention to the corresponding input character during decoding. As depicted in **Fig. 4**, the decoding of subsequent clause with attention mechanism can be expressed as follows:

$$h_t = f(w_s h_{t-1} + w_c C_{2,t-1} + w_v v_t) \tag{3}$$

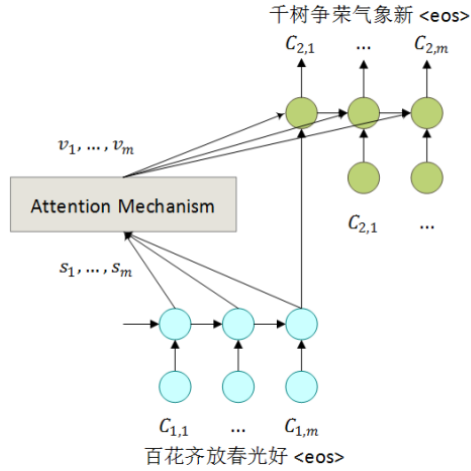

**Fig. 4.** Illustration of the attention-based S2S model.

### 3.3 Processing of Head Character

Based on our proposed three-stage model, the quality of generated complete acrostic couplet heavily depends on the first antecedent clause. Due to its sequential and iterative generation procedure, vanilla RNN based antecedent clause generator requires a reasonable head character (i.e. start token) for high-quality generation. As a part of our design of user interaction (UI), user input is constrained in 4 characters. As an optimal method of improving user experience, Naive Bayes is employed for sampling two characters from user input, which is expressed as:

$$P(B|A) = \frac{P(B)P(A|B)}{P(A)} \tag{4}$$

Where event A represents the appearance frequency of character C while event B represents the frequency of character C appearing in antecedent sentence as head character. Based on train corpus, this Naïve Bayes model can be trained with maximum likelihood loss.

### 3.4 Re-ranking

According to the restrictions of Chinese acrostic couplets, a ranking score is calculated for re-ranking, including length score $s_l$, repeated score $s_r$, tone score $s_t$ and sentiment score $s_s$, which can be denoted as:

$$s = w_l s_l + w_r s_r + w_t s_t + w_s s_s \tag{5}$$

Where weight parameters $w_l$, $w_r$, $w_t$ and $w_s$ are empirically set and optimized. Length score means whether the two clauses of couplet have the same length. Repeated score checks whether repeated characters exist in couplets. Tone score determines whether the ending characters of two clauses exhibit opposed tone. The sentiment score is calculated based on a SWM model, which is higher for positive couplets.

## 4 Experiments and Evaluations

In this section, dataset processing and experimental settings are described at detail. Moreover, re-ranking method for clause candidate selection is introduced. In addition, couplet generation based on LM as baseline models are used.

### 4.1 Dataset

Firstly, a large couplet corpus is collected to efficiently train the generation model, which consists of approximately 602858 couplets. As a result, a primitive vocabulary of 7318 characters was achieved. After omitting low frequency characters less than 10 times, the vocabulary size is decreased to 5647. Additionally, specific symbols are added into the trained vocabulary, including '<unk>' representing low frequency characters and '<eos>' donating the end of sentence. Moreover, 1000 and 2000 couplets are randomly sampled for validation and testing.

### 4.2 Parameter Setting and Training

In this paper, word2vec is used for distributed representation of Chinese characters, which is initially pre-trained with the random initialization (Mikolov et al., 2013). Herein, each character are mapped into a low and dense dimensional vectors, where 256-dimensional word embedding is used. The LSTM cell in antecedent and subsequent sentence generator both have 1000 hidden units. The cell layer number of LM and seq2seq models is 2 and 4, respectively. To ensure generation diversity, group size and width of beam search is set to 4 and 2, respectively. As a result, 16 candidates are achieved according to the constraints of Spring Festival couplet.

All of the trainable parameters are randomly initialized within the range [-0.5, 0.5]. They are trained by stochastic gradient descent to minimize the cross-entropy loss with the Adam optimizer (Kingma and Ba, 2015). The mini-batch size of 128 is chose for training. Moreover, to prevent gradient explosion, the gradient is clipped to the maximum of 5. The learning rate is initially set to 0.001 and adaptively decreased along with training.

### 4.3 Evaluation Metrics

**Automatic Evaluation**

The Bilingual Evaluation Understudy (BLEU) (Papineni et al., 2002) score is widely used for evaluation of machine translation. In this paper, BLEU metric is chosen as an automatic evaluation approach for our couplet generation, where original couplets are used as reference ground truth. Note that BLEU score can represent the similarity between generated couplets and human-written ones. Moreover, some specifications related to Chinese couplets are taken into consideration, such as Length Matching, Character Structure and Tone Pairing. Among them, Length Matching means that both couplet sentence must have the same length. Character Structure ensures that both sentences shouldn't contain the same characters and/or phrases in the same positon. Tone Pairing requires the last characters in the two sentences exhibit opposed tone.

**Human Evaluation**

Different from machine translation, BLEU score is not enough for couplet generation evaluation due to its high diversity. Therefore, eight graduate students majored in traditional Chinese are asked to review generated couplets. They are asked to score our generated couplets with 1-5 scores in three aspects, including Structural Symmetry, Semantic Coherence and Topic Relevance. In terms of Structural Symmetry, correspondence of each character in the same position from two sentences in the aspect of part of speech (POS) and semantics. Semantic Coherence means that antecedent and subsequent clauses are semantically coherent but not repetitive. Fluency examines whether both clauses of generated couplets are expressed fluently.

**Table 1.** Automatic evaluation results.

| Method | Length Matching | Character Structure | Tone Pairing | BLEU |
|---|---|---|---|---|
| LM | 1.0 | 0.60 | 0.86 | 0.2830 |
| SPC Generator | 1.0 | 0.73 | 0.99 | 0.2831 |

### 4.4 Experimental Results

Table 1 depicts automatic evaluation results of our proposed couplet generator compared with the baseline. Apparently, both models can easily generate two same length clauses. However, our proposed model performs better in Character Structure and Tone pairing, which is owing to the encoder-decoder network structure and attention mechanism. Note that both models perform comparatively in terms of BLEU score, verifying the ineffectiveness of BLEU metric for literature creation. Human evaluation results are shown in Table 2. Our proposed model performs better in all three aspects than baseline. Other than attention mechanism, both pipelines including selection of head character and re-ranking of candidate clauses contribute to improving the topic relevance of the generated couplets.

Table 2. Human evaluation results.

| Method | Character Correspondence | Semantic Coherence | Topic Relevance | Average |
|---|---|---|---|---|
| LM | 3.50 | 3.46 | 2.69 | 3.22 |
| SPC Generator | 3.82 | 3.62 | 3.38 | 3.61 |

## 5   Conclusions

In this paper, we propose a novel three-stage neural network pipeline method for Chinese couplet generation. The antecedent generator consists of a LM model equipped with a CBS method after statistical selection of head character using the Naïve Bayes. Afterwards, an attention-based S2S model is trained to generate diverse subsequent clauses which are submitted to the re-ranking pipeline for selection of better results. Both automatic and human evaluations demonstrate better performance of our proposed generation system. Moreover, a mini-program of acrostic couplet generation based on our model has also been developed and deployed on Wechat for real users.

## 6   Acknowledgement

This work was supported by Ping An Technology (Shenzhen) Co., Ltd, China.